\documentclass[12pt, onecolumn]{article}

\usepackage{graphicx}
\usepackage[letterpaper,
            bindingoffset=0in,
            left=1in,
            right=1in,
            top=1in,
            bottom=1in,          footskip=.25in]{geometry}

\usepackage{natbib, url, amsmath, xcolor, color, soul, lineno, microtype, subcaption, comment, amsfonts, graphicx}
\usepackage[onehalfspacing]{setspace}
\begin{document}
\title{Benchmarking Model Predictive Control and Reinforcement Learning Based Control for Legged Robot Locomotion in MuJoCo Simulation}

\date{}

\author{Shivayogi Akki, Tan Chen
\thanks{Shivayogi Akki and Tan Chen are with the Department
of Electrical and Computer Engineering, Michigan Technoogical University, Houghton, MI 49931, USA.
 E-mail: sakki@mtu.edu and tanchen@mtu.edu
 } }


\maketitle

\begin{abstract}
Model Predictive Control (MPC) and Reinforcement Learning (RL) are two prominent strategies for controlling legged robots, each with unique strengths. RL learns control policies through system interaction, adapting to various scenarios, whereas MPC relies on a predefined mathematical model to solve optimization problems in real-time. Despite their widespread use, there is a lack of direct comparative analysis under standardized conditions. This work addresses this gap by benchmarking MPC and RL controllers on a Unitree Go1 quadruped robot within the MuJoCo simulation environment, focusing on a standardized task-straight walking at a constant velocity. Performance is evaluated based on disturbance rejection, energy efficiency, and terrain adaptability. The results show that RL excels in handling disturbances and maintaining energy efficiency but struggles with generalization to new terrains due to its dependence on learned policies tailored to specific environments. In contrast, MPC shows enhanced recovery capabilities from larger perturbations by leveraging its optimization-based approach, allowing for a balanced distribution of control efforts across the robot's joints. The results provide a clear understanding of the advantages and limitations of both RL and MPC, offering insights into selecting an appropriate control strategy for legged robotic applications.
\end{abstract}



\section{Introduction}
In the rapidly evolving field of robotics, legged robots stand out for their ability to navigate unstructured terrains where wheeled robots typically struggle \citep{SpringerHandbookOfRobotics}. Applications such as disaster response and exploration benefit from the adaptability of legged locomotion\citep{siegwart2015legged,JumpingLocomotion_RobotsOnTheMoon}. Early-legged robots featured simple control systems like open-loop feedforward control that permitted limited mobility \citep{AdaptiveLocomotion_MultileggedRobotRoughTerrain}. Over the past few decades, advancements in technology have led to the development of more complex legged robots, featuring multiple joints and limbs that require seamless coordination to maintain balance and achieve locomotion on varied surfaces \citep{DesignPrinciplesQuadrupedsImplementation_MITCheetah, Anymal_DynamicQuadrupedalRobot}. This complexity has necessitated the adoption of advanced control strategies. Initially, predefined gaits and closed-loop feedback mechanisms, such as Proportional-Derivative (PD) controllers, were used to achieve locomotion. However, these traditional controllers lack the adaptability needed for more complex and unpredictable environments \citep{UntetheredQuadrupedalRunning_BoundingGaitTheScoutII}. Consequently, more sophisticated methods have been developed, including Central Pattern Generators (CPGs), Model Predictive Control (MPC), and Reinforcement Learning (RL). 

CPGs generate adaptive and synchronized limb movements by mimicking the rhythmic patterns found in biological systems. By adjusting CPG parameters and integrating with sensory feedback, the coupled oscillator models create stable rhythmic patterns that can be modulated to produce various gaits such as walking, trotting, and galloping \citep{CPGLocomotion_AnimalsAndRobots}. MPC is an advanced control strategy that optimizes future control actions by predicting the system's behavior over a defined horizon. It continuously solves an optimization problem at each time step, incorporating system constraints to determine the most effective control inputs \citep{zak2017introduction}. RL has been used to train locomotion policies on quadrupedal robots, enabling them to adapt to various terrains through interaction with the environment and receiving feedback on their performance \citep{RLRobotics_Survey}.

Given the diverse applications of these control strategies, it is evident that selecting an appropriate control method is crucial for optimizing robot locomotion. Currently, the two most prominent control strategies in robot locomotion are RL and MPC \citep{EfficientRL, Terrain-AwareLocomotion_RLandOC, MPC_EnvironmentAdaptation, FeedbackMPC_TorqueControlledLocomotion}. Despite extensive research in both MPC and RL, there is a lack of a direct comparative analysis evaluating their performance in legged robot locomotion under consistent experimental conditions. This paper provides a benchmarking study comparing MPC and RL, implemented on the Unitree Go1 quadrupedal robot within a controlled simulation environment \citep{unitreego1}. It aims to guide the selection of control strategies for advanced robotic systems, enhancing their functionality and deployment in real-world scenarios. Additionally, identifying the limitations of each method provides valuable insights that can help integrate them with other classical control approaches to improve the performance of legged locomotion.

Because of the structural similarities between Go1 and other quadrupedal robots such as Boston Dynamics' Spot and MIT's Cheetah, the controllers discussed in this paper can be adapted to other robotic platforms, making the work useful to researchers working with diverse legged robots. Note that the MPC requires a model while the RL is entirely model-free. Their performance comparison is inherently complex, as each method offers distinct advantages and limitations. To address this, this paper focuses on a fundamental and standardized locomotion task to highlight the differences and illustrate their characteristics in a simulated environment facilitated by MuJoCo. This approach also ensures a fair evaluation by minimizing the influence of other factors, with a focus on the comparison of control performance. Section \ref{Section_LitReview} provides a review of the work on RL, MPC, and the combined controllers with some examples.  Section \ref{Section_Prob} outlines the problem formulation and the algorithm setup. Section \ref{Section_ExResults} describes the experimental results and further discussion in Section \ref{Section_Discussion}. Finally, Section \ref{Section_Conclusion} concludes this work and suggests directions for future research.


\section{Related Work} \label{Section_LitReview}

This section reviews the work on RL, MPC, and combination methods for legged locomotion. It is worth noting that a direct comparative analysis between RL and MPC under a standardized task and controlled environment has not been conducted. This paper addresses this gap by systematically evaluating the strengths and limitations of RL and MPC controllers for legged locomotion.

\subsection{Reinforcement Learning}

Reinforcement Learning \citep{sutton2018reinforcement} is a machine learning approach where an agent learns to make decisions by interacting with an environment to maximize cumulative rewards. Early RL methods, like Q-learning \citep{watkins1992q}, evolved to Deep Reinforcement Learning \citep{mnih2013playing, mnih2015human} with the integration of neural networks, enabling complex problem-solving. Modern algorithms such as Proximal Policy Optimization (PPO \citep{schulman2017proximal}) and Soft Actor-Critic (SAC \citep{haarnoja2018soft}) address stability and convergence issues, extending RL’s applicability to high-dimensional, continuous environments. RL has been applied to robotics for tasks like locomotion \citep{margolis2024rapid, kimura1997reinforcement}, manipulation \citep{gu2017deep, jeon2023learning}, and navigation in uncertain environments. It has also been used in autonomous vehicles for decision-making and path planning, leveraging its ability to learn from interaction and adapt in real time. 

\subsection{Model Predictive Control}

Model Predictive Control is an advanced control strategy that optimizes control actions by predicting future system behavior over a finite horizon using a mathematical model. It has become popular for controlling multivariable systems with constraints due to its ability to handle dynamics in real time. Some
excellent review articles were published, focusing on both its theoretical aspects \citep{henson1998nonlinear,bemporad2007robust, schwenzer2021review} and its practical applications \citep{qin2003survey}. Recent developments in MPC focus on enhancing computational efficiency, robustness, and adaptability. Aligned with advancements in computational power and the availability of complex process models for various systems, MPC has enabled control over systems that were previously considered unmanageable. Its applications have extended beyond traditional chemical processes to diverse fields, including unmanned aerial vehicles \citep{singh2001trajectory} and legged robotics \citep{farshidian2017real,di2018dynamic,gaertner2021collision}.

\subsection{Combination of RL and Model-based Control}

Recent research has explored various strategies to combine these approaches effectively \citep{ha2024learningbasedleggedlocomotionstate, bellegardaanonline}. One method involves using RL to optimize the parameters of MPC, enhancing its performance and adaptability. For instance, \citep{app13010154} proposed a framework where RL adjusts MPC parameters to improve locomotion performance and stability in quadruped robots. Another approach employs MPC as a supervisory controller to ensure safety and stability, while RL handles tasks requiring adaptability. \citep{chen2024learningagilelocomotionadaptive} combined MPC with RL to achieve agile and robust quadruped locomotion, effectively managing the trade-off between computational complexity and real-time control. These studies underscore the potential of combining RL and MPC to develop control systems that are both robust and adaptable, capable of operating effectively in dynamic and uncertain environments. Some research has also been conducted to address the limitations of RL by combining it with other methods to improve the computational efficiency or performance \citep{tutsoy2017learning, hartmann2024deep}.

\section{Methods} \label{Section_Prob}
The primary objective of this work is to benchmark the performance of MPC and RL controllers in a simulated environment. The benchmarking is conducted on a straight walking task at a constant velocity. The task performance is evaluated based on the disturbance rejection ability, the success of recovery under failure, and the energy efficiency. We utilize the Unitree Go1 robot, which is recognized for its reliability and advanced locomotion capabilities in robotics research \citep{unitreego1}. The simulation is conducted in MuJoCo (Multi-Joint Dynamics with Contact), which is a high-performance physics engine designed for simulating complex dynamic systems, particularly in robotics and biomechanics \citep{Mujoco2012,blanco2024benchmarking, ada2022generalization}. Recent work focuses on integrating MuJoCo with high-level libraries and frameworks like OpenAI Gym and RL Baselines, enhancing its capabilities for end-to-end training pipelines in robotics \citep{wang2022myosim}.

This section elaborates on the methods for comparing the two controllers. It includes the simulation environment setup for both MPC and RL along with a presentation of the adopted algorithms. Data processing and analysis are performed in MATLAB. The data and simulation videos are available on GitHub \citep{github}.

\subsection{RL Algorithm and Setup}
The progression of different RL algorithms from value-based methods to policy optimization is illustrated in \citep{SpinningUp2018}. One notable advancement in this transition is Trust Region Policy Optimization (TRPO), which improves policy gradient methods by ensuring monotonic improvement through constrained optimization \citep{pmlrv37schulman15}. TRPO achieves this by introducing the concept of a trust region, which limits the magnitude of policy updates to prevent overly aggressive changes.

The trust region is defined by a constraint on the Kullback-Leibler (KL) divergence between the current policy and the updated policy. This ensures that updates remain within a predefined region where the approximation used in policy optimization remains valid. By adhering to this constraint, TRPO enhances the stability and reliability of policy gradient methods over episodes, preventing performance degradation caused by abrupt or large changes to the policy. TRPO's main contribution lies in its focus on improving the stability and reliability of policy gradient methods by ensuring that each policy update does not significantly deviate from the previous one, thereby theoretically guaranteeing monotonic improvement. 

The optimization problem for TRPO written in terms of expectation is:
\begin{equation}
\max_{\theta} \mathbb{E}_{s, a \sim \pi_{\text{old}}} \left[ \frac{\pi_{\theta}}{\pi_{\text{old}}} A_{\pi_{\text{old}}} \right],
\end{equation}
subject to:
\begin{equation}
\mathbb{E}_{s \sim \pi_{\text{old}}} \left[ D_{\text{KL}}(\pi_{\text{old}} \Vert \pi_{\theta}) \right] \leq \delta,
\end{equation}
where $\mathbb{E}$ represents the expected value operator, $\pi_{\theta}$ is the policy parameterized by $\theta$, generating action $a$ in state $s$, $A_{\pi_{\text{old}}}$ is the advantage function under the old policy, $D_{\text{KL}}$ is the KL divergence, and $\delta$ is a small positive number dictating the size of the trust region. The advantage function $A_{\pi_{\text{old}}}$ represents the difference between the expected return under policy $\pi_{\theta}$ and the average expected return under the old policy $\pi_{\text{old}}$. 

\begin{figure}[htbp]
\begin{center}
\includegraphics[width=16cm]{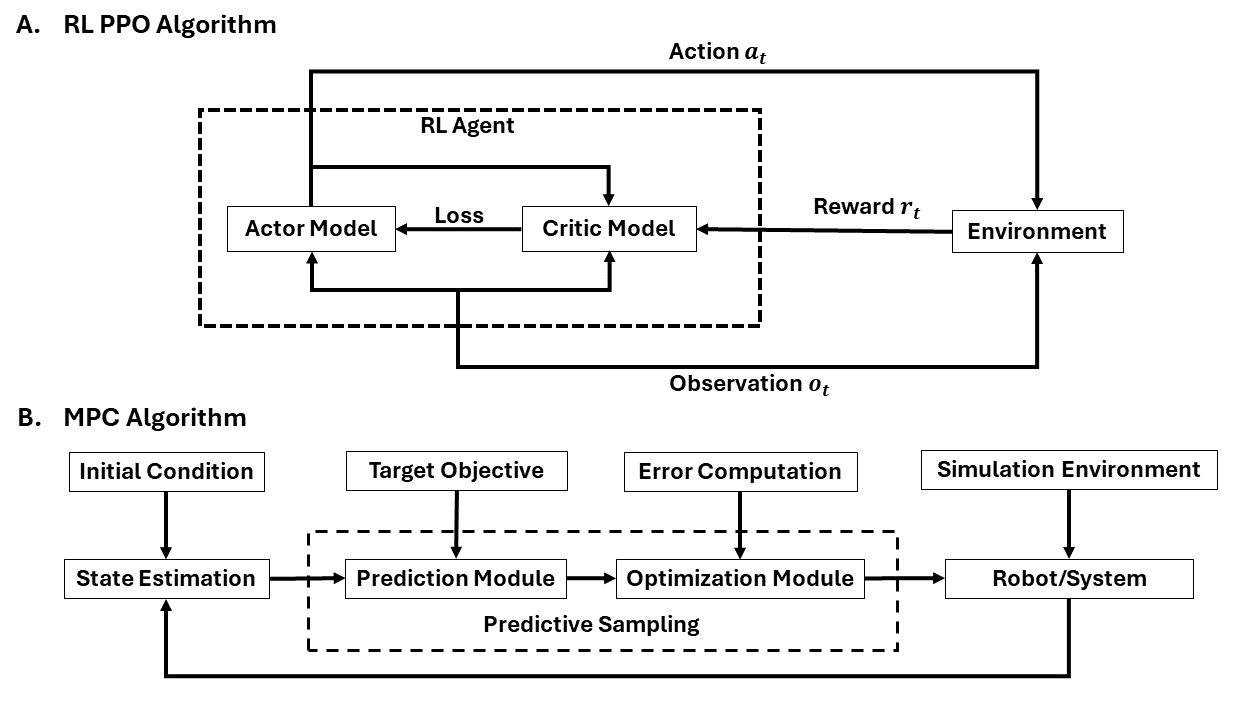}
\end{center}
\caption{\textbf{A.} Reinforcement learning proximal policy optimization algorithm. \textbf{B.} Model predictive control framework using predictive sampling. }
\label{Controller_combo}
\end{figure}

In this research, we adopt an actor-critic type Proximal Policy Optimization (PPO) algorithm for benchmarking (Figure \ref{Controller_combo}-\textbf{A}). Building on TRPO, PPO simplifies TRPO's approach to maintain a trust region by using a clipped objective function, making it easier to implement and scale \citep{schulman2017proximal}. It retains the benefits of TRPO but with less computational complexity. The objective function for PPO is:
\begin{equation}
\begin{split}
L^{\text{CLIP}}(\theta) = \mathbb{E}_{s, a \sim \pi_{\text{old}}} \biggl[ \min ( \frac{\pi_{\theta}}{\pi_{\text{old}}} A_{\pi_{\text{old}}}, \text{clip}( \frac{\pi_{\theta}}{\pi_{\text{old}}}, 1-\epsilon, 1+\epsilon ) A_{\pi_{\text{old}}} )\biggr]
\end{split},
\end{equation}
where $\epsilon$ is a hyperparameter (typically small, e.g., 0.1 or 0.2) that controls the extent to which the policy can change in a single update (the ``clipping'' parameter). This objective encourages improving the policy $\pi_{\theta}$ as long as the ratio between the new and old policies remains within the interval $[1-\epsilon, 1+\epsilon]$, effectively constraining the policy update within a ``proximal'' region around the old policy.

The RL training environment for Go1 is setup with OpenAI Gym \citep{OpenAIgym} and RL Baselines3 zoo (rlzoo3) \citep{rl-zoo3} in the MuJoco physics simulation environment. To obtain hyperparameters and run the RL algorithm, the process begins with optimization using Optuna, an automatic hyperparameter optimization framework \citep{akiba2019optuna}. After optimal hyperparameters are identified, they are used to train the PPO algorithm from the rlzoo3. For the readers' interest, some of the key hyperparameters used for training the locomotion task include setting the policy to MlpPolicy, a learning rate of $5.9e-4$, $128$ steps per update, a total of $1e6$ timesteps, and a clip range of $0.2$.

\subsection{MPC Algorithm and Setup}

Building on the fundamental principles of MPC, the predictive sampling algorithm in the MuJoCo MPC (MJPC) framework can simplify real-time predictive control for complex robotics tasks \citep{howell2022predictive}. In this paper, we utilize the predictive sampling algorithm. The process begins with the system's initial conditions including actuator type and joint states, followed by state estimation as shown in Figure~\ref{Controller_combo}-\textbf{B}. The optimization problem is in the following form:

\begin{equation}
\begin{aligned}
\min_{s_{1:T}, a_{1:T}} & \quad \sum_{t=0}^{T} c(s_t, a_t), 
\end{aligned}
\end{equation}
subject to:
\begin{equation}
\begin{aligned}
& \quad s_{t+1} = m(s_t, a_t), \, \text{given} \, s_0,
\end{aligned}
\end{equation}

where the objective is to optimize state \( s \in \mathbf{S} \) and action \( a \in \mathbf{A} \) trajectories, governed by a model \( m: \mathbf{S} \times \mathbf{A} \rightarrow \mathbf{S} \), which defines the system's dynamics. The minimization goal is to compute the total return, represented as the sum of step-wise values \( c: \mathbf{S} \times \mathbf{A} \rightarrow \mathbb{R}_+ \), along the trajectory from the current state to the horizon \( T \):

\begin{equation} \label{cost_function}
c(s, a) = \sum_{i=1}^{N} w_i \cdot \text{n}_i \big( \mathbf{r}_i(s, a) \big)
\end{equation}

The cost function (in Equation \ref{cost_function}) is expressed as the sum of \( N \) terms, where each term consists of three key components. First, it includes a nonnegative scalar weight \( w \geq 0 \) that defines the relative importance of the term within the overall cost. Second, it incorporates a norm function \( \text{n}(\cdot) \), which operates on a vector and produces a nonnegative scalar value, achieving its minimum when the vector equals zero. Finally, the residual, denoted as $\mathbf{r}(s,a)$, represents a measure of deviation or error in the system's state $s$ and action $a$ relative to the desired task or target. It can include terms like positional errors, velocity errors, or control deviations implemented as custom MuJoCo sensors.

The target objective defines the desired behavior or goal for the system, while the error computation evaluates deviations from this target. Solving this finite-horizon optimal control problem can be achieved through two major classes of algorithms: direct methods and indirect methods \citep{kelly2017introduction}. Predictive sampling is a simple, derivative-free optimization-based direct shooting method used within the MJPC framework. After the target objective defines the desired behavior or goal for the system, while the error computation evaluates deviations from this target, predictive sampling generates multiple control action samples, evaluates their costs, and selects the best path. This process is continuously repeated to adapt to real-time changes (Figure \ref{Controller_combo}-\textbf{B}). Unlike some complex methods requiring detailed mathematical models and gradient calculations, predictive sampling is straightforward and can be used on less powerful hardware. In the MJPC framework, predictive sampling is one of several available planners. 

As MJPC utilizes the MuJoCo physics engine, the non-linearity is handled internally by MuJoCo, and the external disturbances are explicitly modeled within the system dynamics by incorporating additional forces and torques into the equations of motion:
\begin{equation} \label{eqofmotion}
M(q) \ddot{q} + C(q, \dot{q}) \dot{q} + G(q) = \tau + J^\top f_{\text{ext}}, \end{equation}
where $M(q)$ is the mass (inertia) matrix, $\ddot{q}$ represents the generalized accelerations, $C(q,\dot{q})\dot{q}$ captures Coriolis and centrifugal forces dependent on the configuration and velocities, and $G(q)$ represents gravitational forces. The control torques applied by the system's actuators are represented by $\tau$, while $J^{T}$ is the Jacobian transpose mapping external forces $f_{ext}$ from Cartesian space to joint space, and $f_{ext}$ represents the external forces including the contact forces and external disturbances applied to the robot.

In this research, we integrate the Unitree Go1 robot into the extensive pre-existing MJPC package and utilize the predictive sampling method to carry out the benchmarking in the MuJoCo environment as done in RL.
\section{Results}\label{Section_ExResults}

This section presents the simulation experiments and the results of benchmarking MPC and RL controllers. The benchmarking is conducted on a standardized task to ensure that the controllers are evaluated on the same locomotion task, providing an unbiased and consistent basis for comparison.

\subsection{Standardized Task}

\begin{figure}[htbp]
\begin{center}
\includegraphics[width=12cm]{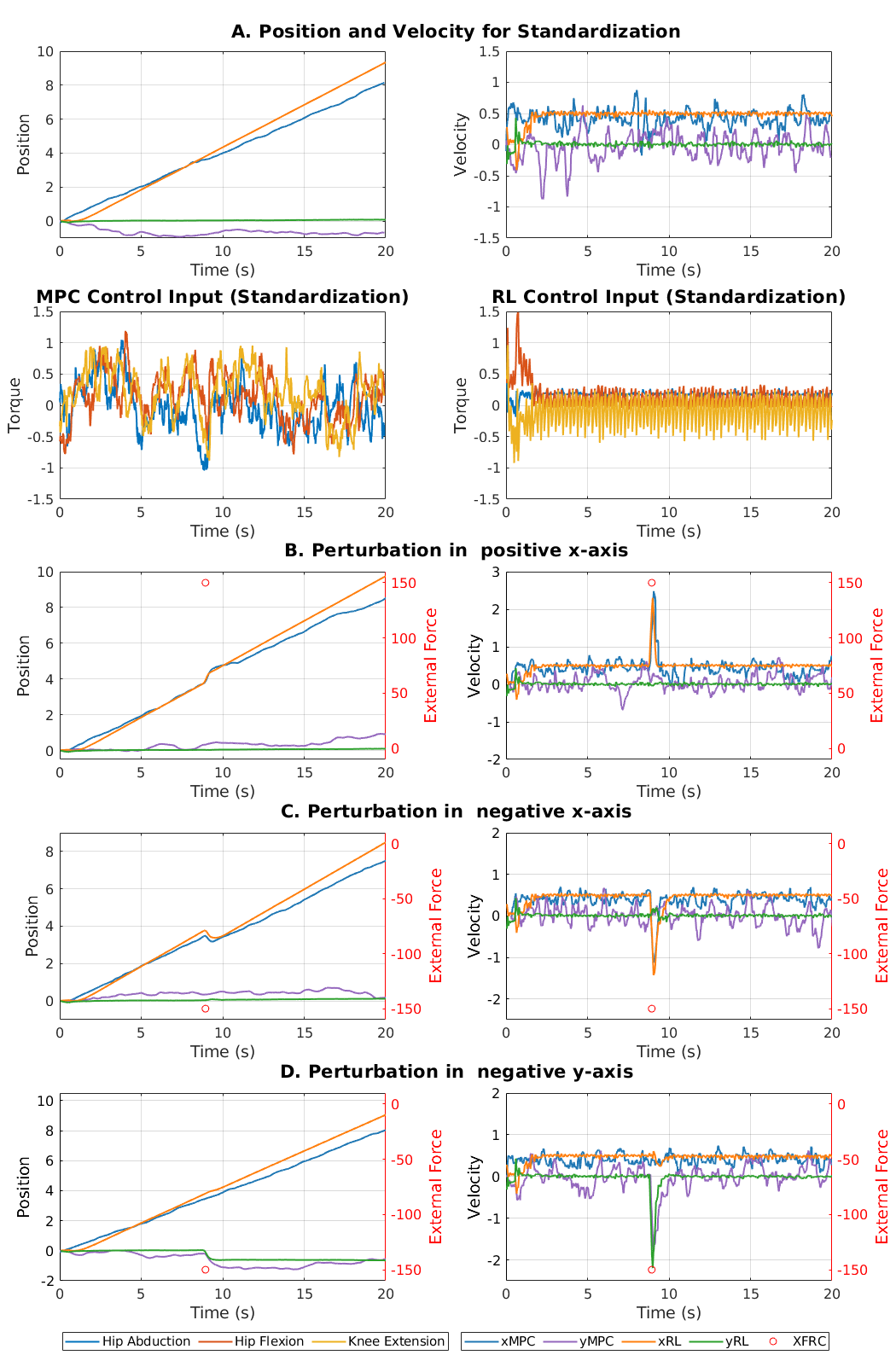}
\end{center}
\caption{\textbf{A.} shows position and velocity plots for the Go1 robot with both MPC and RL controllers during standardized task, and next, the control inputs. \textbf{B.}, \textbf{C.}, and \textbf{D.} illustrate the robot's position and velocity response to perturbations in the positive $x$-axis, negative $x$-axis, and negative $y$-axis, respectively, under both controllers. The graphs demonstrate differences in disturbance rejection capabilities, showing how each controller stabilizes the robot after external forces are applied. The XFRC, marked by the red circle, indicates the perturbation force and timestep when the perturbation is applied. }\label{Perturbation_combo}
\end{figure}

In this paper, the standardized task chosen for the evaluation involves a straightforward locomotion scenario: walking in a straight line along the $x$-axis at a constant velocity of $0.5 m/s$. Figure \ref{Perturbation_combo}-\textbf{A} represents the results of the standardization task, which ensures that the environments set up in both MPC and RL controllers are on equal grounds concerning straight walking at a constant velocity of $0.5 m/s$. This standardization is crucial for justifying further perturbation tests.

The top-left plot shows the position of the center of mass of the Go1 robot over time for both MPC and RL controllers. It demonstrates that the primary motion occurs along the $x$-axis, with both controllers enabling the robot to walk approximately the same distance within a limited time. The top-right graph displays the velocity data over time for both controllers. In RL, the velocities in the $x$ and $y$ directions stabilize around $0.5 m/s$ and $0 m/s$, respectively. This indicates that RL maintains a consistent forward velocity with minimal lateral movement with a variance of $0.0012$. The MPC controller also stabilizes around the same velocities but exhibits fluctuations in the $y$-axis with a variance of $0.056$, suggesting that while MPC can achieve and maintain the desired velocity, it requires adjustments to address these fluctuations.

The bottom plots in Figure \ref{Perturbation_combo}-\textbf{A} show the control inputs for a single limb of the quadruped under both controllers. The three curves correspond to the limb's primary joints: hip abduction, hip flexion, and knee extension (see Figure \ref{FBD_combo}-\textbf{A}). With RL control, it is evident that the knee extension contributes significantly to locomotion, as indicated by its higher amplitude compared to the hip abduction and flexion joints. The higher amplitude indicates that the knee extension joint is predominantly utilized for forward movement. In contrast, the MPC controller achieves locomotion with balanced contributions from all three joints.

\subsection{Disturbance Rejection Capability}
The disturbance rejection capability is evaluated concerning two metrics: the recovery performance after perturbation and the maximum allowable disturbance for each controller. To have a detailed comparison, the disturbances are applied at the center of mass (CoM) along various directions. The recovery performance is evaluated based on the overshoot and settling time after perturbation. See Figure \ref{FBD_combo} for an illustration of the robot and the coordinates.

\subsubsection{Perturbation Tests During Locomotion} \label{Perturbation Tests During Locomotion}

We apply perturbation forces to the robot's CoM in various directions while it maintains a constant velocity of $0.5 m/s$. The timing of the perturbations is randomly selected but remains consistent for both controllers. These perturbations are handled as external forces in Equation (\ref{eqofmotion}). Additionally, perturbation tests with different magnitudes were performed, all yielding similar results. The results presented are from tests with a force of $150 N$, applied for $0.2$ seconds.

\noindent\textbf{A. Perturbation in positive $x$ direction}

Figure \ref{Perturbation_combo}-\textbf{B} illustrates the response of the Go1 robot to the perturbation applied in the positive $x$-axis (forward direction) during a straight walk at $0.5 m/s$, using MPC and RL controllers. The position plots (left column) show that the position under RL control exhibits a smaller deviation (smaller overshoot) of $0.4$ in the $x$-axis than the MPC controller with a deviation of $0.65$ just after perturbation. Both controllers can converge the robot to the predesigned constant velocity. The velocity plots (right column) further show that while both controllers experience disturbances at the point of impact, the RL controller stabilizes approximately $0.25s$ faster (with a shorter settling time), suggesting robustness against perturbation.

In terms of control input, the RL controller significantly relies on the hip flexion joint to compensate for the external force, whereas the MPC controller distributes the effort across all three joints. This distribution indicates that the RL controller focuses on using specific joints to handle disturbances. In contrast, the MPC controller adopts a balanced approach, utilizing multiple joints to manage external forces.

\noindent\textbf{B. Perturbation in negative $x$ direction}

Similarly, the results of a perturbation of $150 N$ along the negative $x$-axis (opposite to the direction of robot locomotion) while walking at a constant velocity of $0.5 m/s$ are shown in  Figure \ref{Perturbation_combo}-\textbf{C}. In the position plots, the RL controller shows a slightly larger deviation in the $x$-axis position than the MPC controller, taking $0.3s$ longer to recover noticed in the velocity plot. This is due to the backward force causing the robot to reach an almost failed state, with its base barely touching the ground. As aforementioned, the RL controller relies primarily on a single joint resulting in a slower recovery. Our simulation further shows that the RL controller heavily relies on the hip flexion joint to compensate for the external force along the $x$ direction.

\noindent\textbf{C. Perturbation in $y$ direction}

The quadrupedal robot's left and right sides are symmetric, so if the gait were perfectly symmetric, the responses to perturbations in the positive and negative $y$-axis would be identical. However, due to slight asymmetry in the gait, the responses are not identical but remain very similar. Figure \ref{Perturbation_combo}-\textbf{D} shows the robot's response to the perturbation in the negative $y$-axis while moving in the positive $x$ direction. In the velocity plots (right column), the velocity under the MPC controller shows a smaller deviation (larger overshoot) in the $y$-axis than the RL controller just after perturbation. Despite this, the robot under RL control demonstrates a $0.33s$ shorter recovery time, which can be observed from both the position and velocity plots. This suggests that the robot under the RL control is robust enough to reject perturbations along the $y$-axis. Similar behavior is also observed for perturbations in the positive $y$-axis due to the robot's symmetry. 

\subsubsection{Maximum Allowable Disturbance}

We define failure as the point at which the robot is unable to maintain balance across its four-foot contact points, resulting in other parts of the robot, such as the torso, touching the ground. This loss of balance leading to the robot body contacting the ground is a failure, irrespective of the recovery. The maximum perturbation values for both the RL and MPC controllers are determined through a systematic process. Increasing external forces were applied to the robot in both the $x$ and $y$ directions until failure occurs. 

\begin{figure}[htbp]
\begin{center}
\includegraphics[width=12cm]{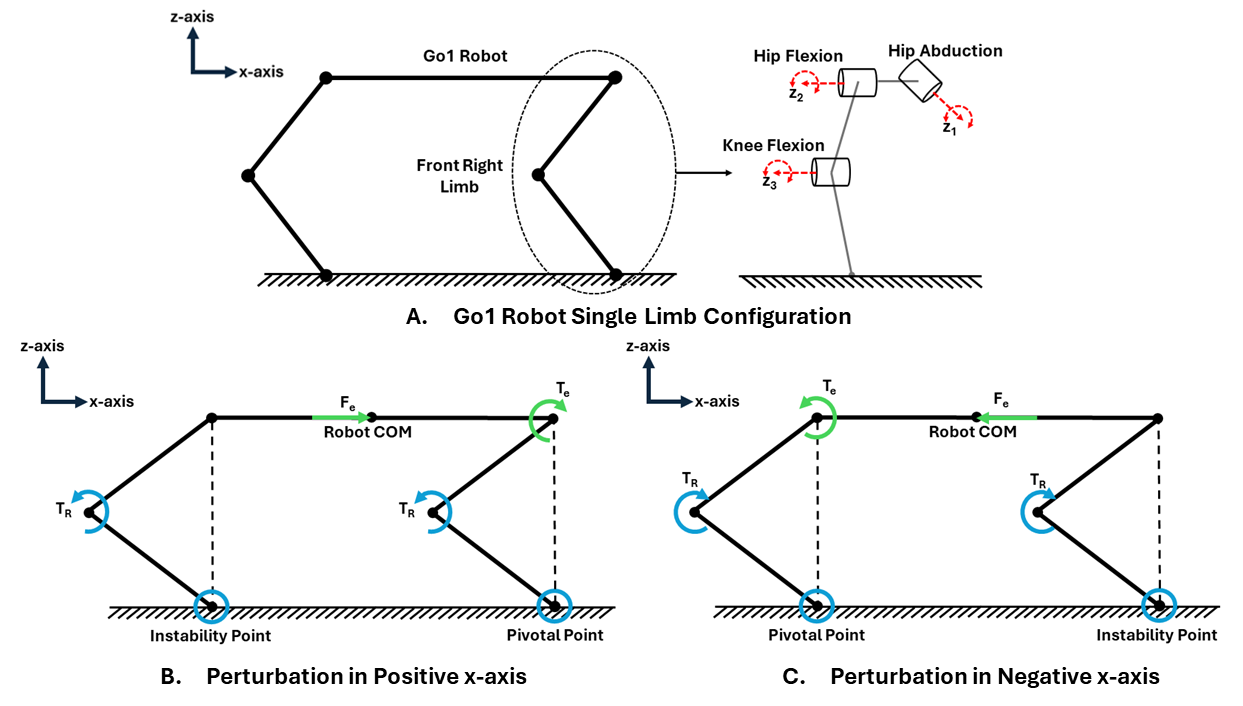}
\end{center}
\caption{\textbf{A.} Go1 Robot Front Right Limb: Illustration of the limb configuration showing hip abduction, hip flexion, and knee flexion joints. \textbf{B.} Perturbation in Positive x-axis: Free body diagram showing the forces and torques on the robot's joints when a perturbation force is applied in the positive x direction of the CoM, leading to instability at the rear foot. \textbf{C.} Perturbation in Negative x-axis: Free body diagram illustrating the forces and torques on the robot's joints when a perturbation force is applied in the negative x direction of the CoM, stabilizing the robot as it pushes down on the pivotal point.}\label{FBD_combo}
\end{figure}

By using the RL control, the Go1 robot can withstand a maximum perturbation force of $520N$ in the positive $x$-axis and $950N$ in the negative $x$-axis. Along the $y$-axis, the RL model handles up to $210N$ in the positive direction and $250N$ in the negative direction, beyond which the robot fails to recover locomotion. When tested with the MPC controller, the robot fails at $260N$ in the positive $x$-axis and $330N$ in the negative $x$-axis. Along the $y$-axis, failure occurs at $290N$ in the positive direction and $340N$ in the negative direction. The MPC controller can reject larger disturbances along the $y$-axis, partly because it controls all three joints in a balanced manner. In contrast, the trained RL control policy relies heavily on a single joint, the knee extension, which contributes to the forward locomotion.

In summary, the RL controller exhibits disturbance rejection capabilities characterized by transient performance metrics such as reduced overshoot and accelerated response following perturbations. It is capable of rejecting disturbances of greater magnitude along the $x$-axis. However, due to the trained RL control policy heavily relying on one joint, the knee extension, it cannot reject as large a disturbance along the $y$-axis as the MPC controller does. 

Also note that for both two controllers, the robot can handle significantly larger disturbance in the negative $x$-axis than the positive direction. This increased tolerance in the negative $x$-axis is due to the robot’s ability to maintain ground contact with all four feet under backward perturbation force. This is explained in Figure \ref{FBD_combo}-B and C, which shows the side view of the Go1 robot with an external force $F_{e}$ acting at the CoM on the torso in positive and negative directions, respectively, generating torque $T_{e}$. The robot counters the perturbation torque with torque $T_{R}$. In Figure \ref{FBD_combo}-B, the torque $T_{R}$ causes the robot's foot to lift off the ground, leading to instability. In contrast, Figure \ref{FBD_combo}-C shows that the torque tends to have the robot push the feet to the ground, thus increasing stability around the pivotal point and maintaining balance.

\subsection{Energy Efficiency}

The energy efficiency of each control strategy is assessed based on the Cost of Transport (CoT) \citep{chen2020robustness}. CoT measures the energy required to move a system relative to its weight and speed, providing a standard metric for energy efficiency in locomotion,

\begin{equation} \label{CoT}
     \text{CoT} = \frac{\text{Power\ Input}}{\text{Total\ Weight} \times \text{Speed}} = \frac{P}{mgv}.
\end{equation}

Here, the power input $P$ is computed as the average power throughout the simulation $T$, 

\begin{equation}
    P = \frac{1}{T} \int_{0}^{T} \sum_{i} \tau_i \dot{\theta} dt,
\end{equation}
were $\tau_i$ is the Torque at the $i^{th}$ timestep and $\dot{\theta}$ is the angular velocity.

The calculations of CoT varies among authors, depending on the assumptions regarding the energy absorption and recovery \citep{ijspeert2014biorobotics, lee2020learning}. When calculating the total power $P$, this work set negative values as zero, which is based on the assumption that the energy recovery is either inefficient or not a significant contributor to the system’s net energy cost. This offers a simpler approach and more interpretable calculation of CoT, particularly in cases where energy recovery is not explicitly modeled. Specifically, the formulation of CoT is shown in Equation (\ref{cot_eq}):

\begin{equation}\label{cot_eq}
\text{CoT} = \frac{\displaystyle \sum_{i=1}^{T} \sum_{j=1}^{n} E_{i,j}}{mgd},
\end{equation}
\begin{equation}
E_{i,j} =
\begin{cases} 
F^{\text{a}}_{i,j} \cdot v_{i,j} \cdot \Delta t, & \text{if } F^{\text{a}}_{i,j} \cdot v_{i,j} > 0, \\ 
0, & \text{otherwise}
\end{cases}
\end{equation}

In Equation (\ref{cot_eq}), $n$ represents the total number of joints in the system. The term $F^{\text{a}}_{i,j}$ refers to the force exerted by the actuator at the $i^{th}$ timestep for $j^{th}$ joint, while the $v_{i,j}$ denotes the angular velocity at the $i^{th}$ timestep for the corresponding joint. The parameter $\Delta t$ is the timestep used in the simulation, $m = 12.0kg$ is the total mass of the robot, and $g = 9.81m/s^2$ is the gravity acceleration. The final computed CoT for RL and MPC are $1.768$ and $2.993$, respectively, measured over a constant time $t = 50 sec$. This indicates that the RL controller, focusing on specific joints, can achieve a lower CoT compared to the MPC controller, suggesting an energy-efficient locomotion strategy.

\section{Discussion}\label{Section_Discussion}
This study highlights significant differences between the MPC and RL controllers in terms of robustness and energy efficiency. The trained RL controller exhibits generally larger disturbance rejection capabilities with reduced energy consumption requirements. In contrast, the MPC controller achieves a balanced distribution of control efforts across joints, which contributes to its ability to recover from disturbances and maintain stable locomotion.

\subsection{Failure and Recovery}

One noteworthy observation is the contrasting failure and recovery mechanisms between the two controllers. The RL controller's reliance on specific joints for locomotion poses a challenge when the robot encounters failure, such as when the torso makes contact with the ground. In these scenarios, our experiments show that the RL controller struggles to recover and continue moving. On the other hand, the MPC controller, which evenly distributes control efforts across all joints, can recover from such failures and resume walking. This distinction underscores the importance of joint utilization strategies in ensuring robust recovery capabilities.

\subsection{RL Generalization} 

In this study, we have further evaluated the generalization \citep{ghosh2021generalization} performance of the RL controller across different terrains, specifically focusing on slippery flat surfaces and uneven terrains. The results reveal significant challenges for the RL controller when encountering environments vastly different from its training setting.

\begin{figure}[htbp]
\begin{center}
\includegraphics[width=16cm]{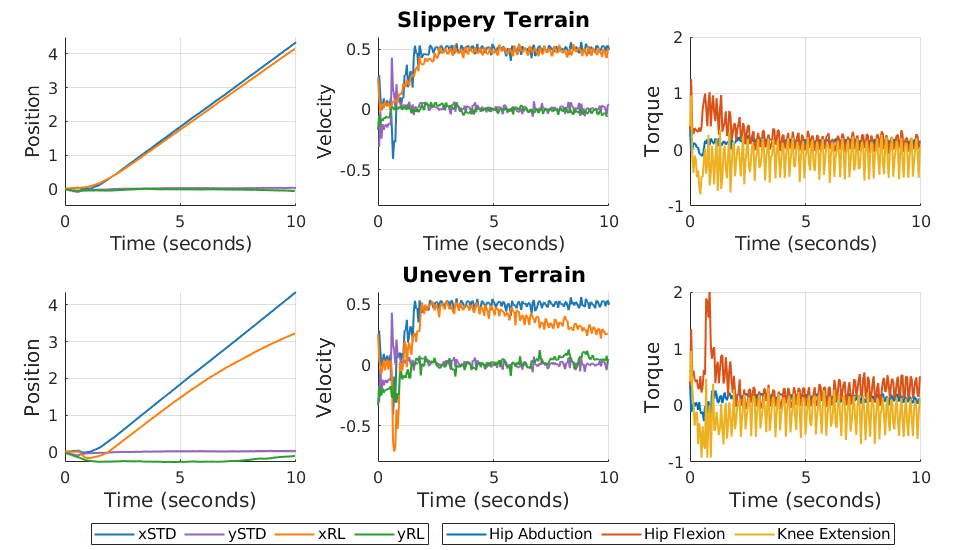}
\end{center}
\caption{The first row shows the Go1 robot's response when controlled by RL on slippery terrain, indicating reduced traction and slower acceleration. The second row presents the robot's response on uneven terrain, demonstrating challenges in maintaining stable movement. Both sets of plots compare position, velocity, and joint torques to illustrate the RL controller's adaptability to varying terrain conditions. The legend shows standardized flat friction terrain (STD) and slippery or uneven friction terrain (RL).}\label{Generalization_combo}
\end{figure}

When testing on a slippery surface, the RL controller, initially trained on flat, frictioned terrain, exhibits a noticeable decrease in the distance traveled within the same time frame (Figure \ref{Generalization_combo}). This reduction occurs primarily due to the compromised movement efficiency from lack of friction. The robot struggles to reach the target velocity of $0.5 m/s$, with a slower acceleration phase reflecting the reduced traction. The controller's control signals are variable, indicating its efforts to adapt to the slippery conditions and maintain stability. Similarly, the performance on uneven terrain demonstrates difficulty in maintaining consistent movement. The robot travels an even shorter distance on the uneven surface (Figure \ref{Generalization_combo}), and the fluctuations in the $x$ and $y$ positions highlight the challenge of sustaining a steady trajectory. The velocity plots also show significant variations in the $x$ and $y$ directions, indicating the robot's struggle to maintain a constant velocity of $0.5 m/s$ amidst varying elevations. The control signals show increased activity in the hip flexion and knee extension joints as the controller attempts to compensate for the terrain irregularities.

These findings show the ability of the RL controller to generalize to new environments, emphasizing the need for specialized training techniques or hybrid approaches to enhance robustness and adaptability. This highlights the importance of continued research into improving the generalization capabilities of RL-based controllers for real-world applications.

\section{Conclusion}\label{Section_Conclusion}

In conclusion, this paper presents a detailed comparison between MPC and RL methods for legged robot locomotion. The results indicate that RL demonstrates larger disturbance rejection, with 0.25s to 0.33s shorter settling time after perturbations, and improved energy efficiency, with a CoT 1.23 lower than that of MPC. This improvement is partly attributed to RL's high-frequency control inputs. In contrast, MPC provides enhanced stability and recovery performance when subjected to large perturbations. Additionally, RL's performance decreases notably on slippery and uneven terrains, highlighting the need for improved generalization capabilities in practical applications. Beyond this comparison, this work makes several key contributions. It establishes robust simulation environments for the Go1 robot in MuJoCo, implementing both RL and MPC approaches. 

Despite promising results in simulation, deep reinforcement learning (DRL) remains highly susceptible to the sim-to-real gap, which arises from the inherent differences between simulated environments and the complexities of the real world \citep{akella2022test, 9981072}. A key direction for future research is validating these approaches in real-world scenarios and adopting strategies to bridge this gap. Techniques such as domain randomization, which introduces variability in simulated parameters like mass, friction, and terrain properties, can improve generalization \citep{peng2018sim,9560769}. Furthermore, system identification methods can calibrate simulation models to closely align with the physical robot's dynamics, thereby minimizing discrepancies between simulation and reality \citep{yu2017preparing}. Another direction for future work involves expanding the fundamental locomotion control task to encompass a broader range of tasks, including turning and obstacle avoidance. Exploring these aspects in future research would further enhance the understanding and development of advanced controllers for legged robots.

\section*{Acknowledgments}
We would like to thank Dr. Zherong Pan for the meaningful discussions and his help in setting up the simulation environments.

\bibliographystyle{Frontiers-Harvard} 
\bibliography{ref}

\end{document}